%% file: main.tex
\documentclass[10pt,twocolumn,letterpaper]{article}

\usepackage{cvpr}              

\usepackage{graphicx}
\usepackage{amsmath}
\usepackage{amssymb}
\usepackage{booktabs}
\usepackage{makecell}
\usepackage{multirow}
\usepackage{overpic}

\usepackage[export]{adjustbox}

\def\eg{\emph{e.g.\ }}
\def\ie{\emph{i.e.\ }}

\def\etal{\emph{et al.\ }}

\usepackage[pagebackref,breaklinks,colorlinks]{hyperref}

\usepackage[capitalize]{cleveref}
\crefname{section}{Sec.}{Secs.}
\Crefname{section}{Section}{Sections}
\Crefname{table}{Table}{Tables}
\crefname{table}{Tab.}{Tabs.}

\usepackage[dvipsnames]{xcolor}
\definecolor{mypink1}{RGB}{255, 0, 108}
\definecolor{mycyan1}{RGB}{39, 255, 235}

\def\cvprPaperID{2502} 
\def\confName{CVPR}
\def\confYear{2023}

\begin{document}

\title{Efficient and Explicit Modelling of Image Hierarchies for Image Restoration}

\author{Yawei Li$^1$ \quad Yuchen Fan$^2$ \quad Xiaoyu Xiang$^2$ \quad Denis Demandolx$^2$ \\ Rakesh Ranjan$^2$ \quad Radu Timofte$^{1,3}$ \quad Luc Van Gool$^{1, 4}$ \\
$^1$Computer Vision Lab, ETH Z\"urich \quad $^2$Meta Reality Labs \quad $^3$University of W\"urzburg \quad $^4$KU Leuven\\
}
\maketitle

\begin{abstract}
   The aim of this paper is to propose a mechanism to efficiently and explicitly model image hierarchies in the global, regional, and local range for image restoration. To achieve that, we start by analyzing two important properties of natural images including cross-scale similarity and anisotropic image features. Inspired by that, we propose the anchored stripe self-attention which achieves a good balance between the space and time complexity of self-attention and the modelling capacity beyond the regional range. Then we propose a new network architecture dubbed GRL to explicitly model image hierarchies in the \underline{G}lobal, \underline{R}egional, and \underline{L}ocal range via anchored stripe self-attention, window self-attention, and channel attention enhanced convolution. Finally, the proposed network is applied to 7 image restoration types, covering both real and synthetic settings. The proposed method sets the new state-of-the-art for several of those. Code will be available at \url{https://github.com/ofsoundof/GRL-Image-Restoration.git}.
\end{abstract}

\section{Introduction}
\label{sec:intro}

Image restoration aims at recovering high-quality images from low-quality ones, resulting from an image degradation processes such as blurring, sub-sampling, noise corruption, and JPEG compression. Image restoration is an ill-posed inverse problem since important content information about the image is missing during the image degradation processes. Thus, in order to recover a high-quality image, the rich information exhibited in the degraded image should be fully exploited.

\begin{figure}[!tb]
    \begin{center}
    \begin{overpic}[width=1.0\linewidth]{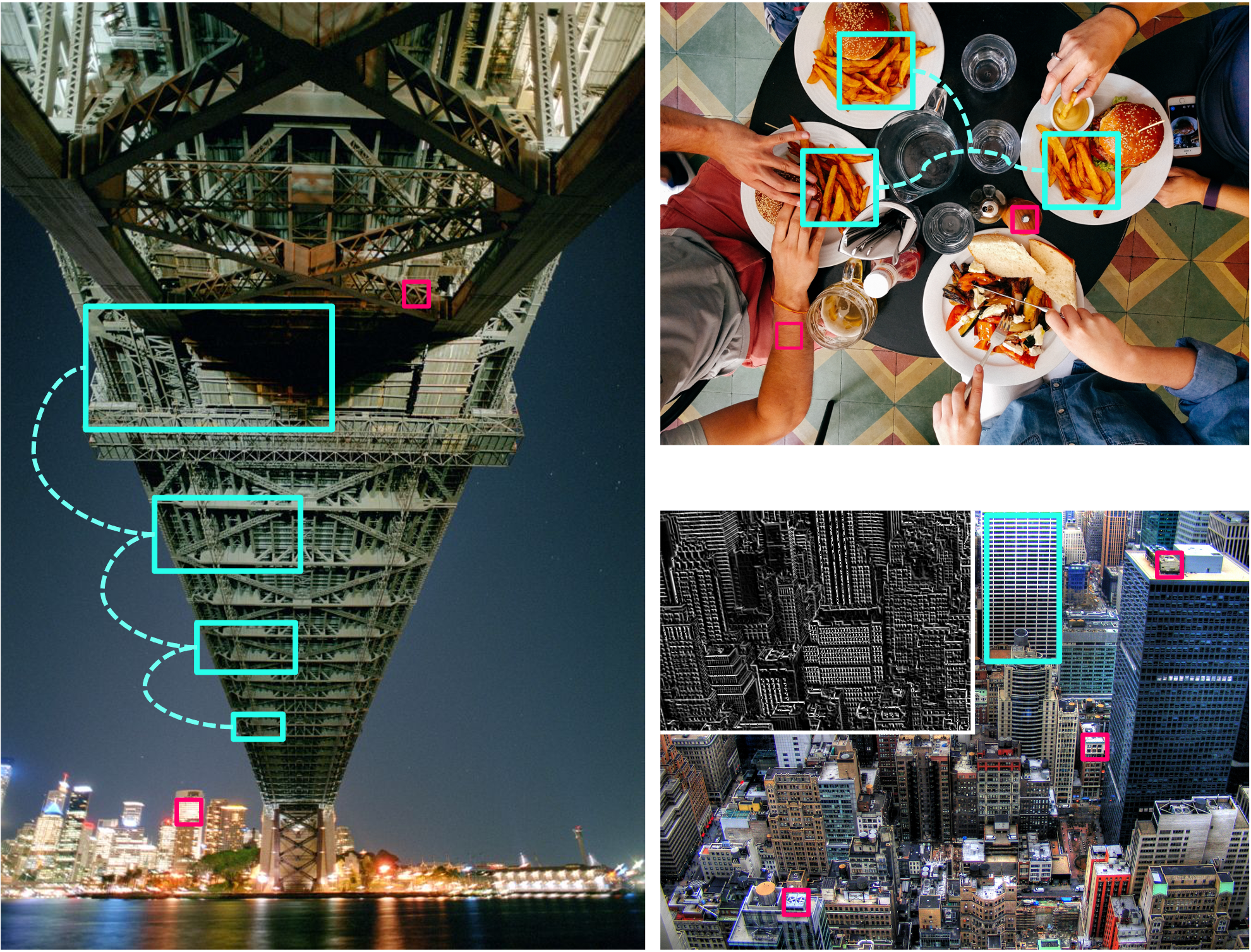}
    \put(0, -3){\scriptsize (a) \textit{bridge} from ICB, $2749 \times 4049$}
    \put(53, 37){\scriptsize (b) \textit{0848x4} from DIV2K, $1020 \times 768$}
    \put(53, -3){\scriptsize (c) \textit{073} from Urban100, $1024 \times 765$}
    \end{overpic}
    \end{center}
    \vspace{-3mm}
    \caption{Natural images show a hierarchy of features in a global, regional, and local range. The local (edges, colors) and regional features (the \textcolor{mypink1}{pink} squares) could be well modelled by CNNs and window self-attention. By contrast, it is difficult to efficiently and explicitly model the rich global features (\textcolor{mycyan1}{cyan} rectangles).}
    \vspace{-6mm}
    \label{fig:motivation}
\end{figure}

Natural images contain a hierarchy of features at global, regional, and local ranges which could be used by deep neural networks for image restoration. 
\textit{First}, the local range covers a span of several pixels and typical features are edges and local colors. To model such local features, convolutional neural networks (CNNs) with small kernels (\eg $3 \times 3$) are utilized. 
\textit{Second}, the regional range is characterized by a window with tens of pixels. This range of pixels can cover small objects and components of large objects (\textcolor{mypink1}{pink} squares in \cref{fig:motivation}). Due to the larger range, modelling the regional features (consistency, similarity) explicitly with large-kernel CNNs would be inefficient in both parameters and computation. Instead, transformers with a window attention mechanism are well suited for this task.
\textit{Third}, beyond local and regional, some features have a global span (\textcolor{mycyan1}{cyan} rectangles in \cref{fig:motivation}), incl. but not limited to symmetry, multi-scale pattern repetition (\cref{fig:motivation}a), same scale texture similarity (\cref{fig:motivation}b), and structural similarity and consistency in large objects and content (\cref{fig:motivation}c). To model features at this range, global image understanding is needed.

\begin{figure*}
    \centering
    \includegraphics[width=\linewidth]{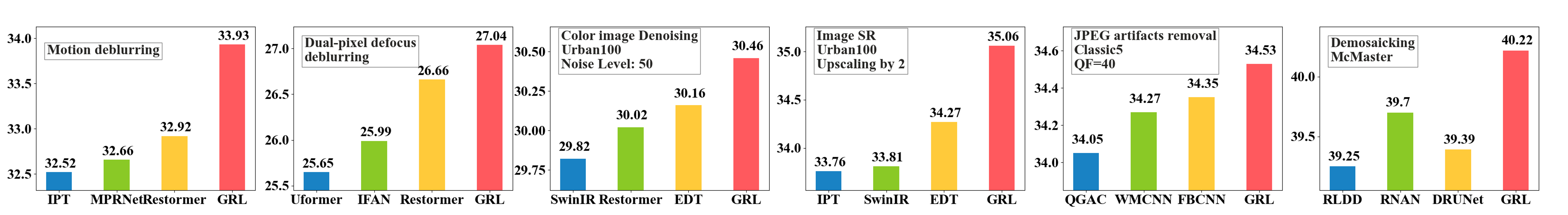}
    \vspace{-6mm}
    \caption{The proposed GRL achieves state-of-the-art performances on various image restoration tasks. Details provided in Sec.~\ref{sec:experiments}.}
    \label{fig:performance_teaser}
    \vspace{-6mm}
\end{figure*}

Different from the local and regional range features, there are two major challenges to model the global range features. Firstly, existing image restoration networks based on convolutions and window attention could not capture long-range dependencies explicitly by using a single computational module. Although non-local operations are used in some works, they are either used sparsely in the network or applied to small image crops. Thus, global image understanding still mainly happens via progressive propagation of features through repeated computational modules. 
Secondly, the increasing resolution of today's images poses a challenge for long-range dependency modelling. High image resolution leads to a computational burden associated with pairwise pixel comparisons and similarity searches. 

The aforementioned discussion leads to a series of research questions: 1) how to efficiently model global range features in high-dimensional images for image restoration; 2) how to model image hierarchies (local, regional, global) explicitly by a single computational module for high-dimensional image restoration; 3) and how can this joint modelling lead to a uniform performance improvement for different image restoration tasks. 
The paper tries to answer these questions in \cref{sec:motivation}, \cref{sec:modelling_image_hierarchy}, and \cref{sec:experiments}, resp.

\textit{First}, we propose anchored stripe self-attention for efficient dependency modelling beyond the regional range. The proposed self-attention is inspired by two properties of natural images including cross-scale similarity and anisotropic image features. Cross-scale similarity means that structures in a natural image are replicated at different scales. Inspired by that, we propose to use anchors as an intermediate to approximate the exact attention map between queries and keys in self-attention. Since the anchors summarize image information into a lower-dimensional space, the space and time complexity of self-attention can be
significantly reduced. In addition, based on the observation of anisotropic image features, we propose to conduct anchored self-attention within vertical and horizontal stripes. Due to the anisotropic shrinkage of the attention range, a further reduction of complexity is achieved. And the combination of axial stripes also ensures a global view of the image content. When equipped with the stripe shift operation, the four stripe self-attention modes (horizontal, vertical, shifted horizontal, shifted vertical) achieves a good balance between computational complexity and the capacity of global range dependency modelling. Furthermore, the proposed anchored stripe self-attention is analyzed from the perspective of low-rankness and similarity propagation. 

\textit{Secondly}, a new transformer network is proposed to explicitly model global, regional, and local range dependencies in a single computational module. The hierarchical modelling of images is achieved by the parallel computation of the proposed anchored stripe self-attention, window self-attention, and channel-attention enhanced convolution. And the transformer architecture is dubbed \textbf{GRL}. 

\textit{Thirdly}, the proposed GRL transformer is applied to various image restoration tasks. Those tasks could be classified into three settings based on the availability of data including real image restoration, synthetic image restoration, and data synthesis based real image restoration. In total, seven tasks are explored for the proposed network including image super-resolution, image denoising, JPEG compression artifacts removal, demosaicking, real image super-resolution, single image motion deblurring, and defocus deblurring. As shown in \cref{fig:performance_teaser}, the proposed network shows promising results on the investigated tasks.

\section{Related Works}
\label{sec:related_works}

\textbf{Convolution for local range modelling.} One of the basic assumptions for example and learning-based image restoration is that repetitive patterns could exist in either the same or different images~\cite{freeman2002example} and that the redundant information they carry could help to restore the local patches. Thus, it helps if repetitive patterns could be detected and modelled~\cite{dong2014learning,kim2016accurate,shi2016real,lim2017enhanced}. This intuition matches the computational procedure of convolution well, which slides the kernel across the image and detects local patterns similar to the learnable kernels. By stacking multiple convolutional layers, the receptive field of a CNN gets progressively enlarged and rich image features are captured. Since the advent of deep learning, great efforts have been made to design CNNs for image restoration~\cite{lai2017deep,zhang2018residual,haris2018deep,wang2018esrgan,wang2018recovering}. 

\textbf{Non-local and global priors.} Besides the local features, it is also important to model the non-local and global image priors. The early work of non-local means serves this purpose, which computes an output pixel as the weighted sum of all the pixels within the image~\cite{buades2005non}. Inspired by that, later works have been developed to utilize the repetitive patterns in a non-local range for image denoising~\cite{dabov2007image} and super-resolution~\cite{glasner2009super}. Apart from the traditional methods, non-local operations are also introduced into deep neural networks for video classification~\cite{wang2018non} and image SR~\cite{liu2018non,zhang2019residual}.

Besides the non-local operations, self-attention has been developed to model the global range dependencies~\cite{vaswani2017attention,devlin2018bert}. 
However, the computational complexity of global self-attention grows quadratically with the number of tokens. Thus, the increase in efficiency of global self-attention is investigated by several works~\cite{child2019generating,katharopoulos2020transformers,choromanski2020rethinking,kitaev2020reformer,wang2020linformer}.

\textbf{Regional self-attention.} Among the methods for accelerating transformers, regional self-attention appears to be promising. The idea is proposed in the pioneering works~\cite{ramachandran2019stand,parmar2018image} and improved as shifted window attention~\cite{liu2021swin,liu2022swin}. Inspired by the success of shifted window attention for visual recognition and perception, this method is also used for image restoration~\cite{liang2021swinir,li2021efficient,chen2022activating}. Despite the good performance of the window attention mechanism, it is pointed out in recent works that a wider range of pixel involvement could lead to better image restoration~\cite{gu2021interpreting,chen2022activating}. 
Thus, in this paper, we try to propose a method that efficiently brings the modelling capacity of self-attention beyond the regional range.

\section{Motivation}
\label{sec:motivation}

\subsection{Self-attention for dependency modelling}

Self-attention is good at modelling long-range dependencies explicitly and it facilitates the propagation of information across the modelled dependencies. 
This operation allows a token to be compared with all the other tokens. The output token is computed as a weighted sum of all the tokens based on a similarity comparison, 
\ie, 
\begin{equation}
    \mathbf{Y} = \mathrm{Softmax}\left( \mathbf{Q} \cdot \mathbf{K}^T / \sqrt{d} \right) \cdot \mathbf{V}, \label{eqn:self_attention}
\end{equation}
where $\mathbf{Q}=\mathbf{X} \cdot \mathbf{W}_{Q}$, $\mathbf{K} = \mathbf{X} \cdot \mathbf{W}_{K}$, $\mathbf{V} = \mathbf{X} \cdot \mathbf{W}_{V}$, $\mathbf{W}_{Q}, \mathbf{W}_{K}, \mathbf{W}_{V} \in \mathbb{R}^{d \times d}$, and $\mathbf{X}, \mathbf{Y} \in \mathbb{R}^{N \times d}$. $N$ and $d$ denote the number of tokens and the dimension of one token, respectively. 
Additionally, $\mathbf{M}$ denotes the attention map, \ie $\mathbf{M} = \mathrm{Softmax}(\mathbf{Q} \cdot \mathbf{K}^T / \sqrt{d})$.

The time complexity of self-attention is $\mathcal{O}(N^2d)$ and the space complexity is dominated by the term $\mathcal{O}(N^2)$ of the attention map $\mathbf{M}$. The computational complexity and memory footprint of self-attention grow quadratically with the number of tokens. Thus, self-attention can easily become a computation bottleneck for images where the number of tokens is the multiplication of the two dimensions of the feature map. To overcome this problem, it is proposed to apply self-attention within a window. In this way, the number of tokens that participate in self-attention is significantly reduced and the computational burden is also lifted.

The problem of window self-attention is that the modelling capacity of the operation is limited to a regional range due to the small window size ($8 \times 8$~\cite{liang2021swinir}). On the other hand, it is shown in recent works~\cite{gu2021interpreting,chen2022activating} that even a slight increase in window size can lead to better image restoration. Thus, it can be conjectured that modelling dependencies beyond the regional range is still important for image restoration. Hence, it remained to be investigated how to maintain the ability for long-range dependency modelling under a controlled computational budget.

\subsection{Motivation I: cross-scale similarity}

\begin{figure}[!tb]
    \centering
    \includegraphics[width=0.98\linewidth]{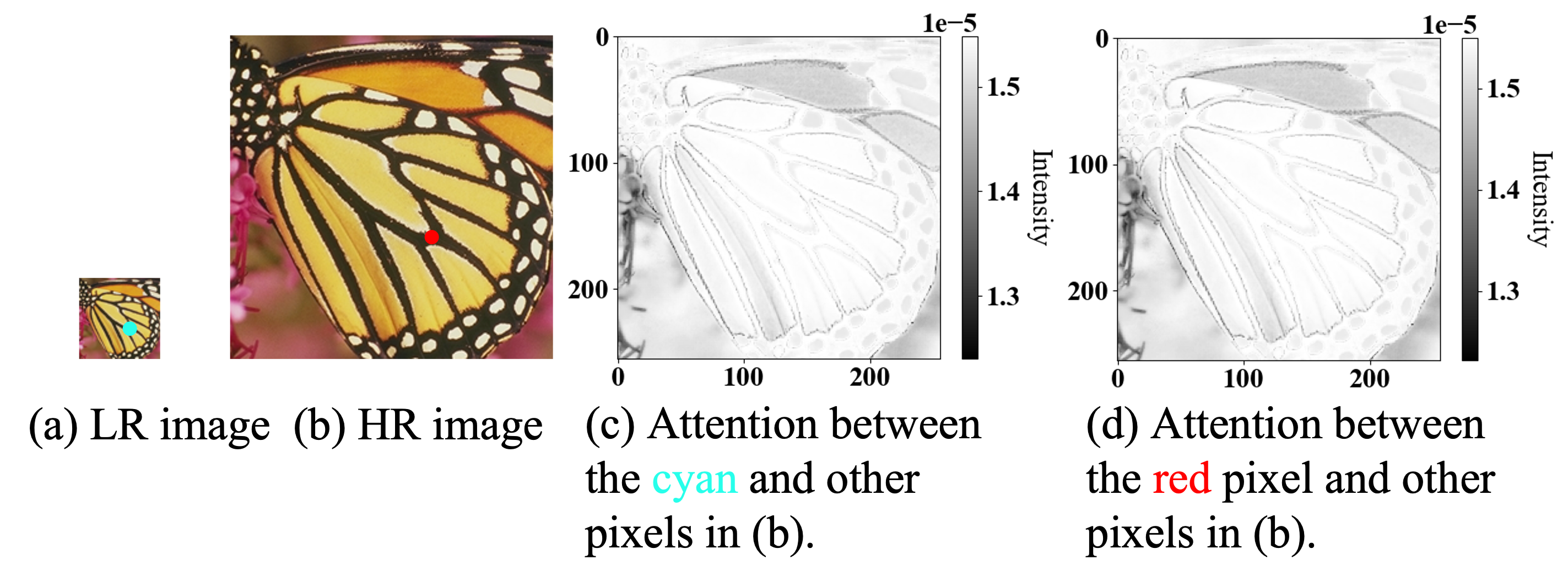}
        \vspace{-6pt}
    \caption{Cross-scale similarity. (c) and (d) shows the attention map between the selected pixels and the example high-resolution image. Although the cyan pixel in (a) and the red pixel in (b) are from images with different resolutions, their attention map with respect to the high-resolution image shows very similar structures.}
    \vspace{-10pt}
    \label{fig:motivation1}
\end{figure}

The attention map $\mathbf{M}$ plays an essential role in self-attention as it captures the similarity between every paired pixels in the image. Thus, improving the efficiency of the self-attention in \cref{eqn:self_attention} needs one to analyze the property of the attention map. And we are inspired by a property of images, \ie cross-scale similarity. That is, the basic structure such as lines and edges of an image is kept in the different versions of the image with different scaling factors. In \cref{fig:motivation1}, the attention map between pixels in an image is shown. Particularly, the attention map between a pixel and the whole image is visualized as a gray-scale heat map. As shown, no matter whether the pixel comes from the high-resolution image or the down-scaled version, the heat map between the pixel and the high-resolution image shows the basic structure of the image. And the heat maps in \cref{fig:motivation1}(c) and \cref{fig:motivation1}(d) are very similar to each other.

\textbf{Anchored self-attention.} Inspired by the cross-scale similarity shown in \cref{fig:motivation1}, 
we try to reduce the complexity of the global self-attention in \cref{eqn:self_attention} by operating on images with different resolutions and manipulating the number of tokens, \ie the $N^2$ term in $\mathcal{O}(N^2d)$. To achieve that, we introduce an additional concept named anchors besides the triplets of queries, keys, and values. The set of anchors is a summary of the information in the image feature map and has a lower dimensionality. Instead of conducting the similarity comparison between the queries and keys directly, the anchors act as an intermediate for the similarity comparison. Formally, the anchored self-attention is proposed as in the following equation
\begin{align}
    \mathbf{Y} & = \mathbf{M}_e \cdot \mathbf{Z} = \mathbf{M}_e \cdot \left( \mathbf{M}_d \cdot \mathbf{V} \right), \label{eqn:anchored_self_attention} \\
    \mathbf{M}_d & = \mathrm{Softmax}( {\mathbf{A} \cdot \mathbf{K}^T} / \sqrt{d}), \\
    \mathbf{M}_e & = \mathrm{Softmax}( \mathbf{Q} \cdot \mathbf{A}^T / \sqrt{d}), 
\end{align}
where $M \ll N$, $\mathbf{A} \in \mathbb{R}^{M \times d}$ is the anchor, $\mathbf{M}_e \in \mathcal{R}^{N \times M}$ and $\mathbf{M}_d \in \mathcal{R}^{M \times N}$ denotes the attention map between the query-anchor pair and anchor-key pair. The choice of the operations to derive the anchors is investigated in the implementation details of the ablation study of the paper.

Since the number of anchors is much smaller than the number of the other tokens, the size of the resulting two attention maps $\mathbf{M}_e$ and $\mathbf{M}_d$ are much smaller than the original attention map $\mathbf{M}$ in \cref{eqn:self_attention}. Then the matrix multiplication in \cref{eqn:anchored_self_attention} is computed from the right hand. The self-attention is first done for the anchors and keys. The attention map $\mathbf{M}_d$ distills the tokens $\mathbf{V}$ into an intermediate feature $\mathbf{Z}$. Then the self-attention is done between the queries and the anchors. The second attention map $\mathbf{M}_e$ expands the size of the feature $\mathbf{Z}$ and recovers the information in $\mathbf{V}$.
The computational complexity of the anchored self-attention is reduced to $\mathcal{O}(NMd)$. And the space complexity is reduced to $\mathcal{O}(NM)$.

\subsection{Motivation II: anisotropic image features}

\begin{figure}[!tb]
\begin{center}
\scriptsize
\begin{tabular}[b]{c@{ } c@{ }  c@{ } c@{ }}
\hspace{-2mm}  
    \includegraphics[width=0.23\linewidth,valign=t]{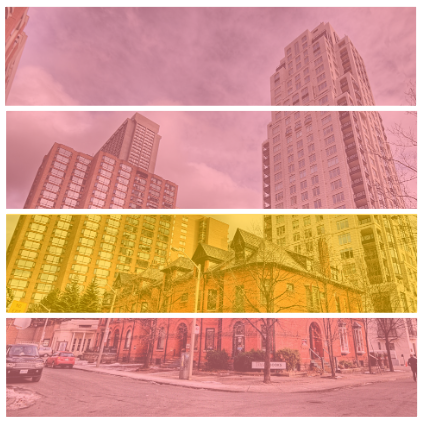} &
    \includegraphics[width=0.23\linewidth,valign=t]{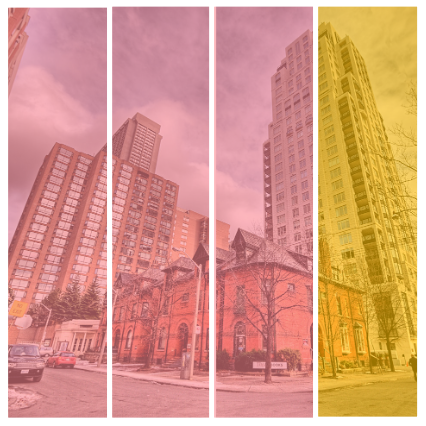} &
    \includegraphics[width=0.23\linewidth,valign=t]{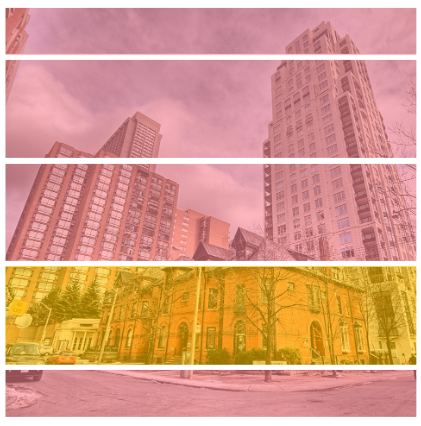} &
    \includegraphics[width=0.23\linewidth,valign=t]{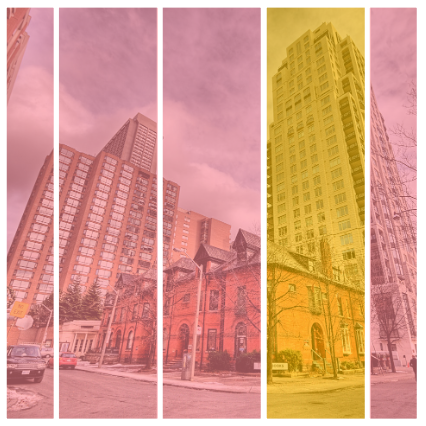}\\

    (a) & (b) & (c) & (d) \\
\hspace{-2mm}  
    \includegraphics[width=0.23\linewidth,valign=t]{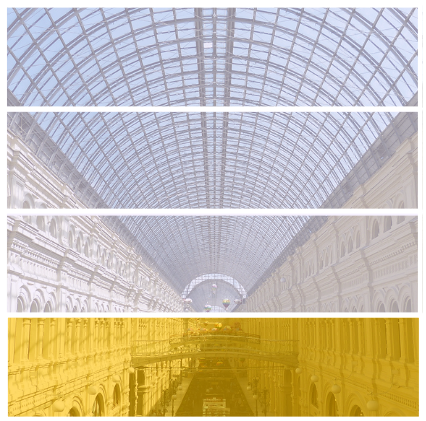} &
    \includegraphics[width=0.23\linewidth,valign=t]{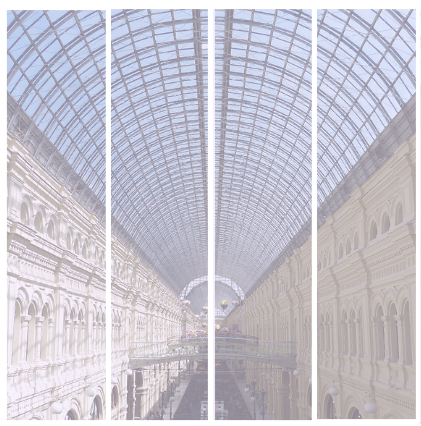} &
    \includegraphics[width=0.23\linewidth,valign=t]{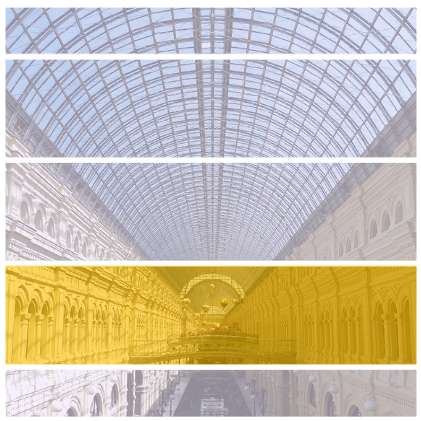} &
    \includegraphics[width=0.23\linewidth,valign=t]{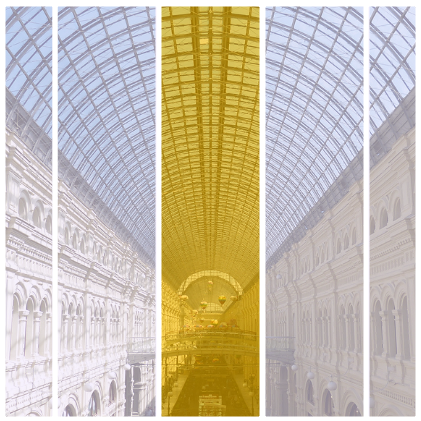}\\
    
    (e) & (f) & (g) & (h)  \\
\end{tabular}
\end{center}
\vspace{-6mm}
\caption{The image features in natural images are anisotropic. Thus, it is not always necessary to employ the uniform global range attention in all parts of the image.}
\label{fig:motivation2}
\vspace{-6mm}
\end{figure}

The anchored self-attention could reduce the space and time complexity of the self-attention in \cref{eqn:self_attention} significantly by removing the quadratic term $N^2$. Yet, for image restoration tasks, the remaining term $N$ is the multiplication of the width and height of the image. Thus, the complexity of the anchored self-attention in \cref{eqn:anchored_self_attention} could still be unaffordable due to the large term $N$. Thus, it is desirable to further reduce the complexity of the anchored self-attention. 

To achieve that goal, we resort to another characteristic of natural images, \ie, the anisotropic image features. As shown in \cref{fig:motivation2}, the natural image features such as the single object in \cref{fig:motivation2}(c)\&(d), the multi-scale similarity in \cref{fig:motivation2}(h), symmetry in \cref{fig:motivation2}(e)\&(g) span in an anisotropic manner. Thus, isotropic global range attention across the entire image is redundant to capture the anisotropic image features. And in response to that, we propose to conduct attention within the anisotropic stripes shown in \cref{fig:motivation2}.

\begin{figure}[!tb]
    \centering
    \includegraphics[width=0.9\linewidth]{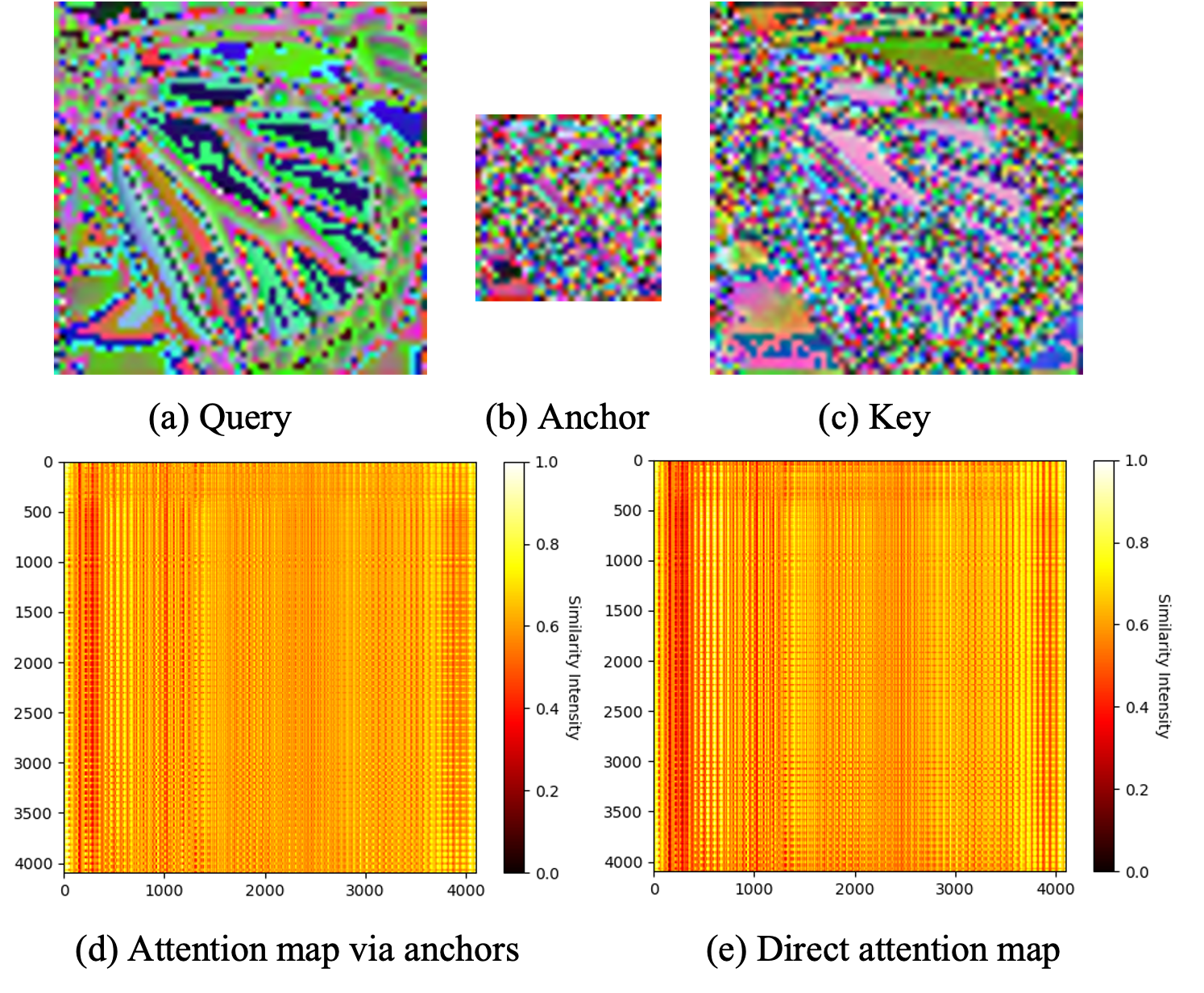}
    \vspace{-4mm}
    \caption{The visualization of the (a) queries, (b) anchors, and (c) keys from the different layers of the proposed network. (d) shows the attention map approximated by \cref{eqn:anchored_self_attention}, \ie $\mathbf{M}_e \cdot \mathbf{M}_d$. (e) shows the exact attention map $\mathbf{M}$ computed in \cref{eqn:self_attention}.}
    \label{fig:attention_map_visualization}
    \vspace{-6mm}
\end{figure}

\textbf{Stripe attention mechanism}. The proposed stripe attention mechanism consists of four modes including the horizontal stripe, the vertical stripe, the shifted horizontal stripe, and the shifted vertical stripe. The horizontal and vertical stripe attention mechanisms could be employed alternately across a transformer network. In this way, a trade-off is made between maintaining the global range modelling capacity and controlling the computation complexity of global self-attention. Thus, in combination with the concept of anchors, we propose the \textbf{anchored stripe self-attention}. For this attention mechanism, efficient self-attention is conducted inside the vertical and horizontal stripes with the help of the introduced anchors.

\begin{figure*}
    \centering
    \includegraphics[width=0.96\linewidth]{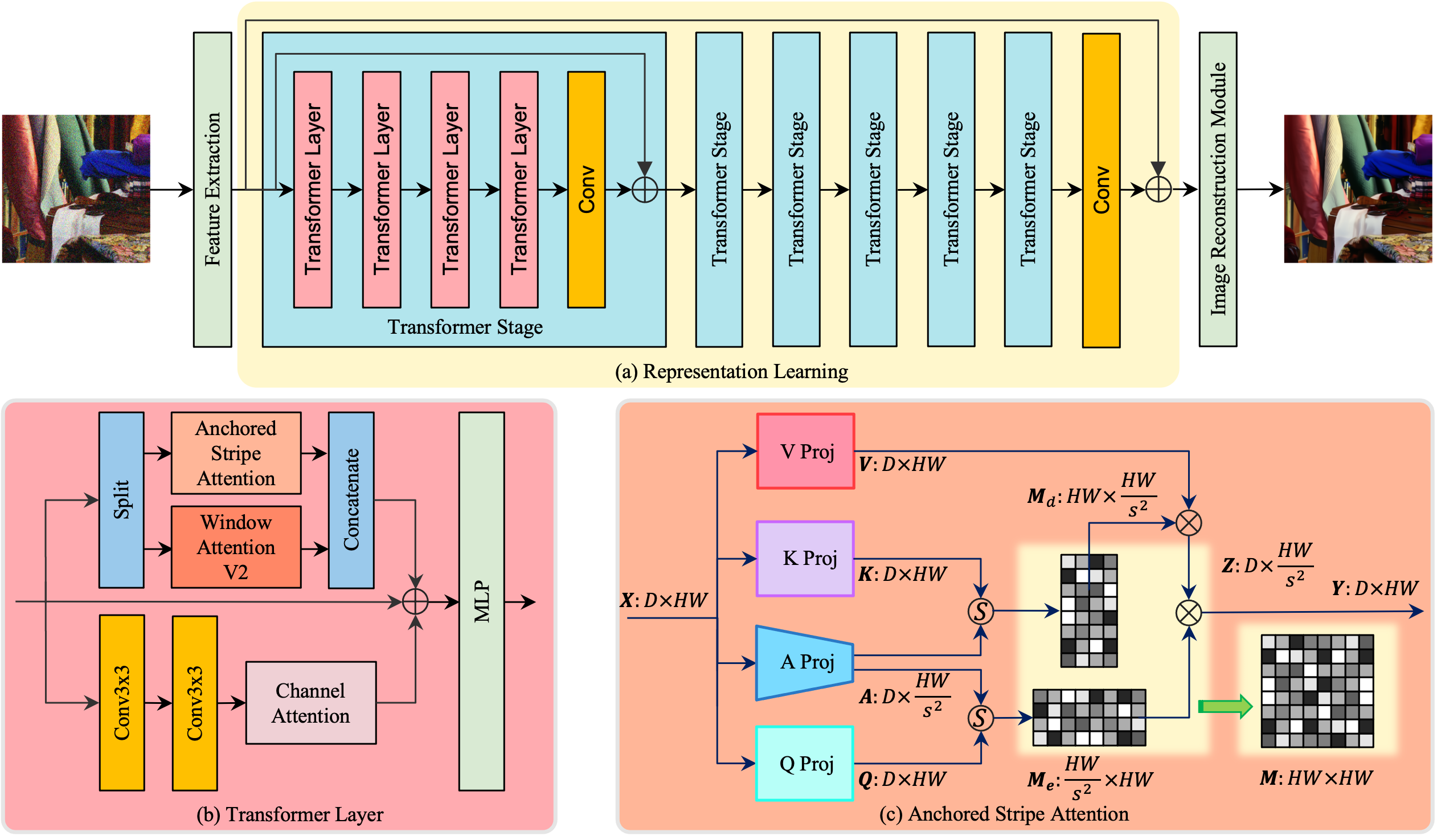}
    \vspace{-4mm}
    \caption{Network architecture. (a) The representation learning module contains stages of transformer layers. (b) The transformer layer is equipped with global, regional, and local modelling blocks. (c) The anchored stripe attention helps to attend beyond regional ranges.}
    \label{fig:network}
    \vspace{-6mm}
\end{figure*}

\subsection{Discussion}
\label{sec:discussion}
The proposed anchored stripe self-attention mechanism is closely related to two other concepts including low-rankness and similarity propagation. And we detail the relationship in this subsection as follows.

\textbf{Low-rankness of attention map.} By comparing the self-attention mechanisms in \cref{eqn:self_attention} and \cref{eqn:anchored_self_attention}, we can easily found out that the original attention map $\mathbf{M}$ is decomposed into small attention maps $\mathbf{M}_d$ and $\mathbf{M}_e$ whose rank is no larger than $M$. And the essence here is to provide the low-rank approximation without calculating the original attention map first. For the success of the anchored self-attention, it is important to ensure that with the anchors as the intermediate, the approximated attention map is similar to the original attention map. Thus, an additional analysis is provided in \cref{fig:attention_map_visualization}. 

First, by observing the queries, anchors, and keys, we can conclude that the anchors have a very similar structure to the query and key. Thus, the anchors are a good summary of the information in the queries and keys. And approximating self-attention with anchors as intermediate seems to be plausible. Additionally, the approximate attention map $\mathbf{M}_e \cdot \mathbf{M}_d$ and the exact attention map $\mathbf{M}$ are also compared in \cref{fig:attention_map_visualization}. As shown, the approximate attention map keeps the major structure in the exact attention map, which is confirmed by the large Pearson correlation coefficients (0.9505) between the two attention maps. Thus, the quality of the anchored self-attention is guaranteed.

\textbf{Metric and similarity propagation.} From another perspective, in the proposed anchored self-attention, the queries and keys are first compared with the anchors and then the query-key similarity is computed. Thus, this computation procedure needs to propagate the query-anchor and key-anchor similarity to the query-key pair. And similarity propagation is related to the triangle inequality in a metric space~\cite{hsieh2017collaborative,wang2004multi,gairola2020simpropnet}.
A mathematical metric needs to satisfy several conditions including the essential triangle inequality, $d(\mathbf{q}, \mathbf{k}) \leq d(\mathbf{a}, \mathbf{q}) + d(\mathbf{a}, \mathbf{k})$, 
where $d(\cdot, \cdot)$ defines a metric between two entities. Thus, the $\mathbf{q}$ / $\mathbf{k}$ distance is upper-bounded by the sum of the $\mathbf{a}$ / $\mathbf{q}$ distance and the $\mathbf{a}$ / $\mathbf{k}$ distance. This implies that if $\mathbf{a}$ is similar (close) to both $\mathbf{q}$ and $\mathbf{k}$, then $\mathbf{q}$ and $\mathbf{k}$ should also be similar (close) to each other. Yet, the similarity measure in \cref{eqn:self_attention} and \cref{eqn:anchored_self_attention} is defined by the dot product instead of the distance between tokens, which does not satisfy the triangle inequality. Thus, similarity propagation could not be theoretically guaranteed. To study the influence of the similarity measure, an ablation study is conducted and the results are shown in \cref{sec:experiments}. Dot product and distance are compared as a similarity measure. According to the results, although the dot product does not strictly obey the triangle inequality, it still guarantees better image restoration results. Thus, we can conclude empirically that the dot product is enough for similarity propagation.

\section{Modelling Image Hierarchies}
\label{sec:modelling_image_hierarchy}

In this section, we answer the second research question described in the introduction, that is, how to explicitly model image hierarchies by a single computational module. In response to that, we propose the GRL network architecture that incorporates \underline{g}lobal range, \underline{r}egional range, and \underline{l}ocal range image modelling capacities.

\textbf{Network architecture.} The overall architecture of the proposed network is shown in \cref{fig:network}. The network takes a degraded low-quality image as input, processes the image inside the network, and outputs a recovered high-quality image. In detail, the network contains three parts. 1) The feature extraction layer is implemented as a simple convolution and converts the input image into feature maps. 2) The representation learning component enriches the information extracted in the previous operation.
The transformer stage consists of several transformer layers and ends with a convolution layer. The dimension of the feature map is maintained across the whole representation learning module. Skip connection is applied to both the transformer stage and the representation learning module. 3) The image reconstruction module takes the rich features calculated by the previous operations and estimates a recovered image.

\input{tables/defocus_deblur_table.tex}

\input{tables/motion_deblur_table.tex}

\textbf{Transformer Layer.} This layer in \cref{fig:network}b is the key component that provides the hierarchical image modelling capacity in the global, regional, and local range. This layer first processes the input feature map by the parallel self-attention module and channel attention enhanced convolutions. The convolution branch serves to capture local structures in the input feature map. On the other hand, the self-attention module contains the window attention proposed in Swin transformer V2~\cite{liu2022swin} and the anchored stripe attention proposed in this paper. The feature map is split equally along the channel dimension and concatenated along the channel dimension again after the parallel processing within the two attention modules. The window attention provides the mechanism to capture the regional range dependencies. Then the feature maps outputted by the convolution module and the attention module are added to the input feature map, which is processed by the following MLP module.

\input{tables/motion_deblur_realblur.tex}
\input{tables/dn_table.tex}

\input{tables/sr_table.tex}

\textbf{Anchored stripe self-attention.} The operation of the proposed anchored stripe attention is conducted according to \cref{eqn:anchored_self_attention} and visualized in \cref{fig:network}c. The dimension of different features is also shown. The triplet of $\mathbf{Q}$, $\mathbf{K}$, $\mathbf{V}$ is derived by plain linear projections. To summarize the information into anchors, the anchor projection is implemented as an average pooling layer followed by a linear projection. After the anchor projection, the resolution of the image feature map is down-scaled by a factor of $s$ along both directions. As shown in \cref{fig:network}, the two attention maps $\mathbf{M}_d$ and $\mathbf{M}_e$ play a similar role as the original attention map $\mathbf{M}$ but with less space and time complexity.

\section{Experimental Results}
\label{sec:experiments}

The experimental results are shown in this section. We answer the third research question raised in the introduction by investigating the performance of the proposed network on different image restoration tasks. Based on the data type, the investigated tasks are classified into three commonly used settings including 1) real image restoration (single-image motion deblurring, defocus deblurring), 2) image restoration based on synthetic data (image denoising, single image SR, JPEG compression artifact removal, demosaicking), and 3) real image restoration based on data synthesis. We provide three networks with different model sizes including the tiny, small, and base versions (GRL-T, GRL-S, GRL-B). For real and synthetic image restoration, Adam optimizer and $L_1$ loss are used to train the network with an initial learning rate $2\times 10^{-4}$. More details about the training dataset, training protocols, and additional visual results are shown in the \textit{supplementary material}.

\input{tables/jpeg_table.tex}
\input{tables/demosaicking_table.tex}

\subsection{Image deblurring}

We first investigate the performance of the proposed network on two real image restoration tasks including single-image motion deblurring, and motion deblurring.

\noindent \textbf{Single image motion deblurring}. \cref{table:motion_deblurring} and \cref{table:motion_deblurring_realblur} shows the experimental results for single image motion deblurring on synthetic datasets ({GoPro} \cite{nah2017deep}, {HIDE} \cite{shen2019human}) and real dataset ({RealBlur-R}~\cite{rim2020real}), respectively. Compared with the previous state-of-the-art Restormer~\cite{zamir2022restormer}, the proposed GRL achieves significant PSNR improvement of 1.01 dB on the GoPro dataset. On the HIDE dataset, the PSNR improvement is 0.43 dB. Please note that the improvement is achieved under fewer parameter budget. As shown in \cref{table:denoising}, GRL-B saves 24\% parameters compared with Restormer. As shown in \cref{table:motion_deblurring_realblur}, GRL-B sets the new state-of-the-art performance of 40.20 PSNR on RealBlur-R dataset.

\noindent \textbf{Defocus deblurring}. \cref{table:defocus_deblurring} shows the experimental results for defocus deblurring using single image and dual-pixel images. Our GRL outperforms the previous methods for all three scene types. Compared with Restormer on the combined scenes, our GRL achieves an elegant performance boost of 0.20 dB and 0.38 dB for single and dual-pixel defocus deblurring. Compared with Uformer~\cite{wang2022uformer} and IFAN~\cite{Lee_2021_CVPRifan}, GRL achieves PSNR gain of 1.39 dB and 1.05 dB for the dual-pixel setting.  

\subsection{Image restoration based on synthetic data}
Investigating image restoration with synthetic data is also valuable to reveal the network capacity of restoration methods. Besides the experiments on the real data, we also study the performance of the network on synthetic data.

\noindent \textbf{Image denoising}. First, the experimental results on Gaussian image denoising are shown in \cref{table:denoising}. For a fair comparison between different models, both the network complexity and accuracy are shown in the table. And several key findings are observed. \textbf{I.} The tiny version GRL-T is extremely efficient, reducing model complexity by two orders of magnitude (only 0.76\% of ~\cite{chen2021pre} and 2.7\% of DRUNet~\cite{zhang2021plug}) while not sacrificing network accuracy. \textbf{II.} The small version GRL-S performs competitive with the previous state-of-the-art SwinIR~\cite{liang2021swinir} and Restormer~\cite{zamir2022restormer}. \textbf{II.} On Urban100, the base version outperforms Restormer by a large margin (\eg 0.44dB PSNR gain for color image and noise level 50).

\noindent \textbf{Image SR}. Experimental results for classical images are shown in \cref{tab:sr_results}. Both lightweight models and accurate SR models are summarized. A similar conclusion could be drawn from the results. \textbf{I.} Among the lightweight networks, GRL-T outperforms both convolution and self-attention-based networks including DBPN~\cite{haris2018DBPN}, SwinIR~\cite{liang2021swinir} and EDT~\cite{li2021efficient}. Compared with EDT, Significant improvements are obtained on Urban100 and Manga109 datasets (0.44 dB and 0.22 dB for $\times 4$ SR). \textbf{II.} GRL-B sets the new state-of-the-art for accurate image SR. \textbf{III.} GRL-S achieves a good balance between network complexity and SR image quality. 

\noindent \textbf{JPEG compression artifact removal}. The experimental results for color and grayscale images are shown in \cref{table:jpeg_car}. Four image quality factors ranging from 10 to 40 for JPEG compression are studied. As shown in the table, the proposed GRL-S network outperforms the previous state-of-the-art method elegantly across different datasets and quality factors. Notably, GRL-S has a much smaller model complexity than FBCNN.

\noindent \textbf{Demosaicking}. Results for image demosaicking is shown in \cref{table:demosaicking}. The proposed method outperforms the previous methods RNAN~\cite{zhang2019residual} and DRUNet~\cite{zhang2021plug} significantly.

\subsection{Real image restoration based on data synthesis}
\begin{figure}[!tb]
\begin{center}
\scriptsize
\begin{tabular}[b]{c@{ } c@{ } c@{ } c@{ }}
\hspace{-2mm}  
    \includegraphics[width=0.24\linewidth,valign=t]{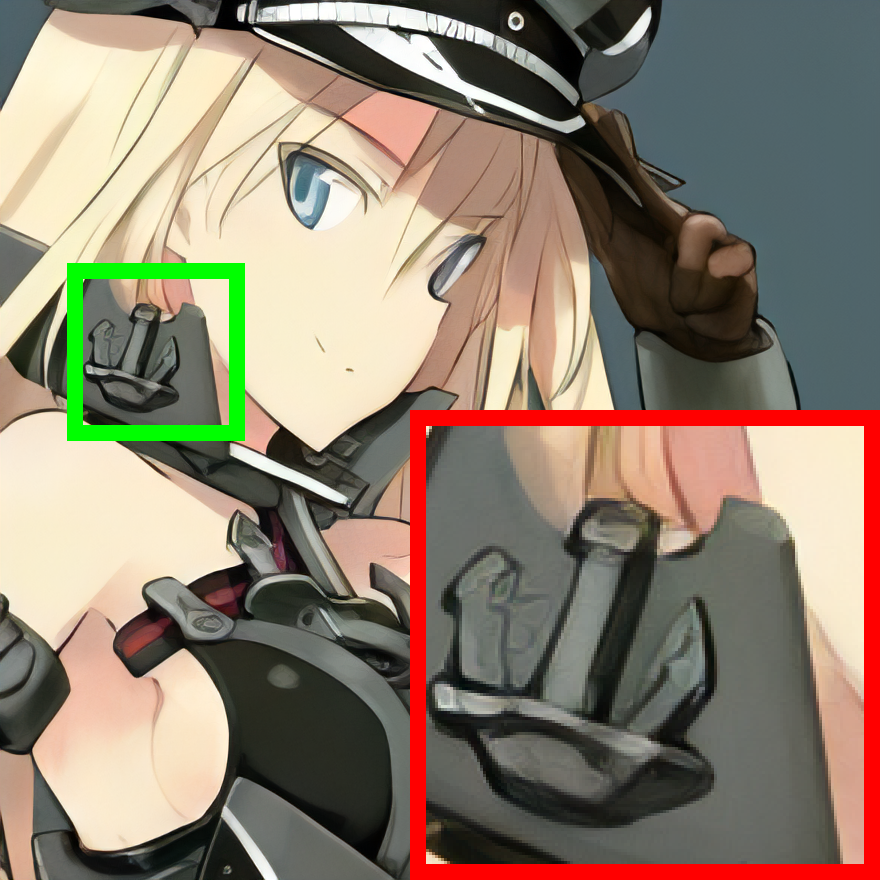} &
    \includegraphics[width=0.24\linewidth,valign=t]{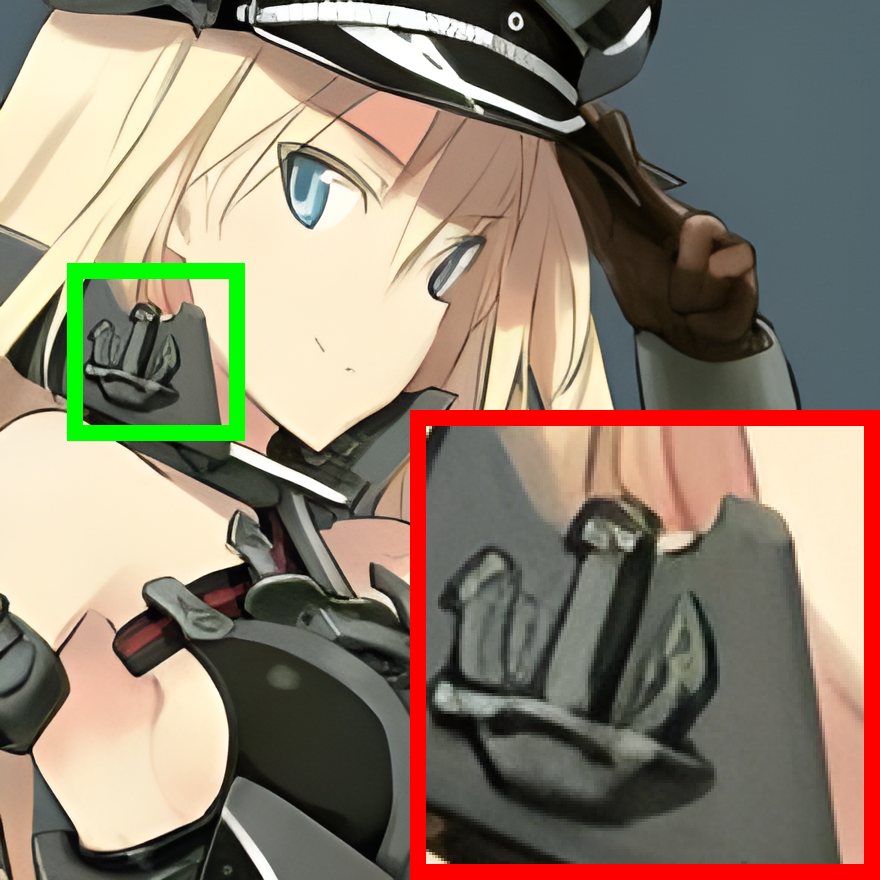} & 
    \includegraphics[width=0.24\linewidth,valign=t]{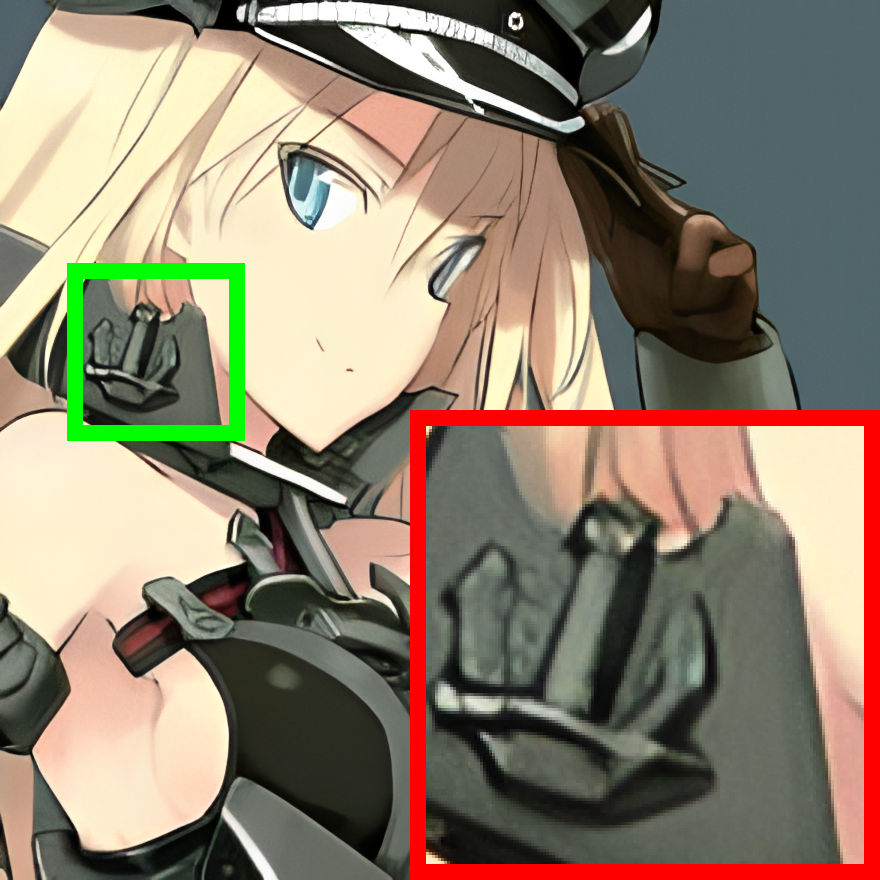} &
    \includegraphics[width=0.24\linewidth,valign=t]{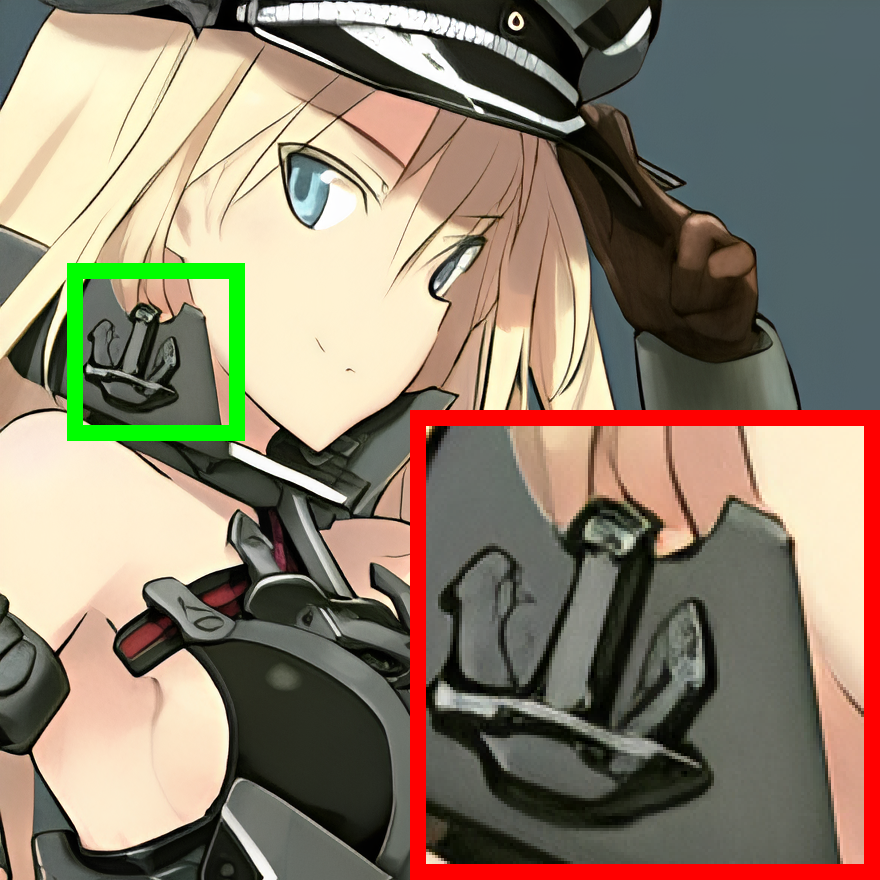} \\
\hspace{-2mm}  
     \includegraphics[width=0.24\linewidth,valign=t]{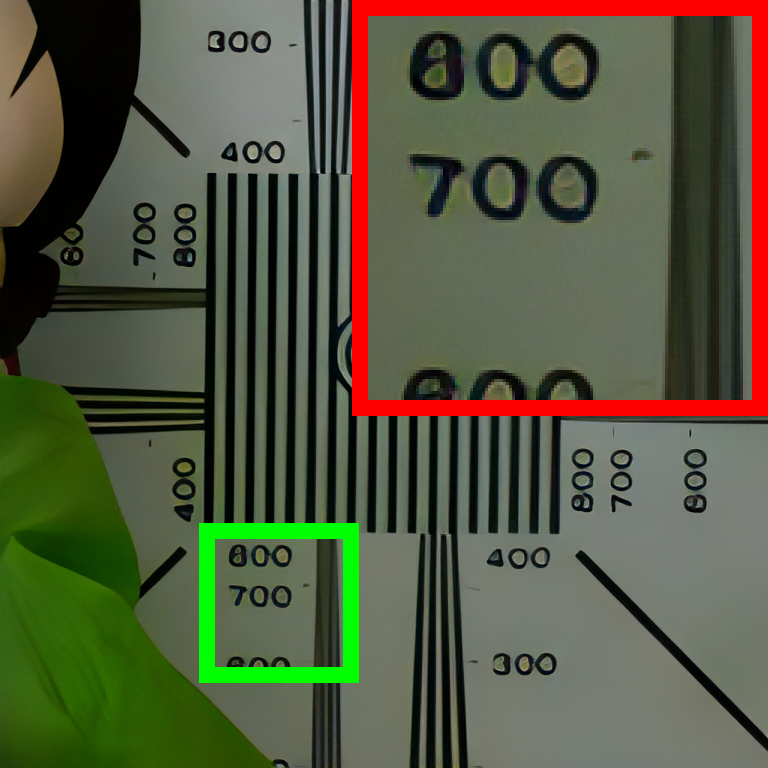} &
    \includegraphics[width=0.24\linewidth,valign=t]{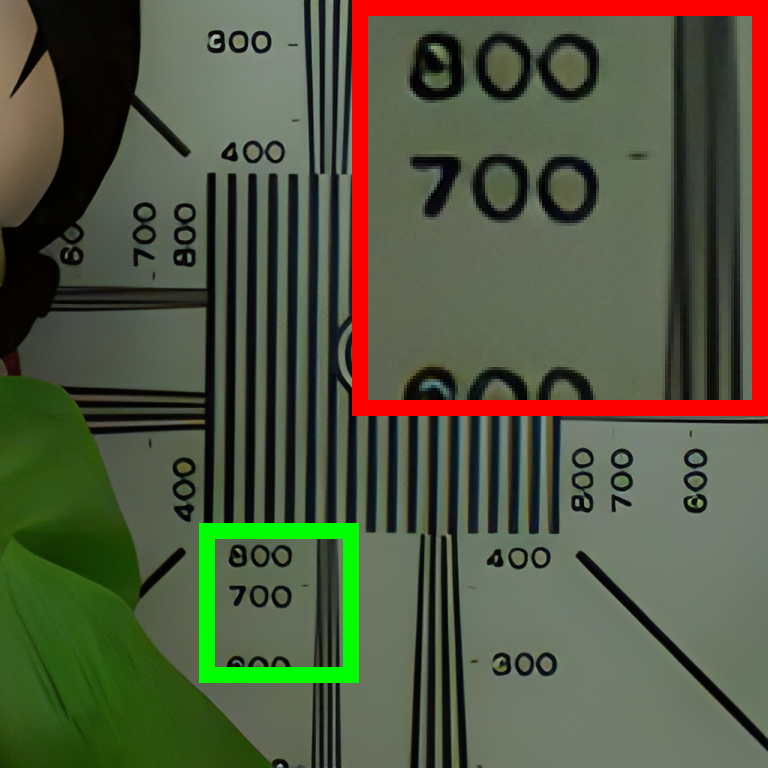} & 
    \includegraphics[width=0.24\linewidth,valign=t]{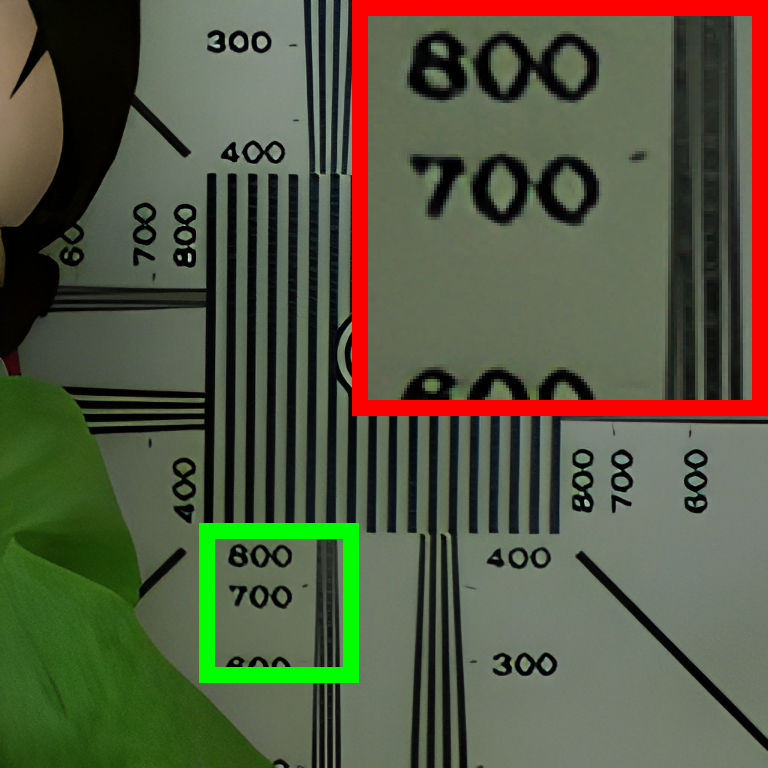} &
    \includegraphics[width=0.24\linewidth,valign=t]{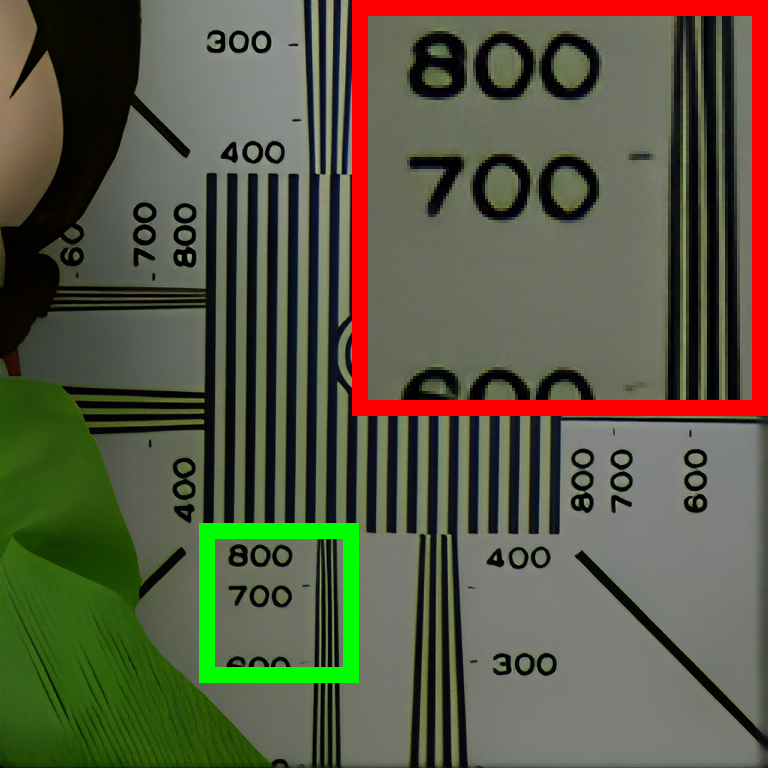} \\
    BSRGAN & Real ESRGAN & SwinIR & GRL  \\
\end{tabular}
\end{center}
\vspace{-4mm}
\caption{Visual results for real-world image SR.}
\label{fig:realsr}
\vspace{-6mm}
\end{figure}

Finally, we also investigate the performance of the network for real-world image restoration. The aim is to super-resolve a low-quality image by an upscaling factor of 4. Since there are no ground-truth images for this task, only the visual comparison is given in \cref{fig:realsr}. Compared with the other methods, the proposed GRL is able to remove more artifacts in the low-resolution images.

\subsection{Ablation study}

\noindent \textbf{Influence of the similarity comparison method}. As mentioned in \cref{sec:discussion}, for theoretical guarantee of similarity propagation, a mathematical metric rather than a dot product should be used. To study the difference between, image restoration with the two operations are compared and the results are shown in \cref{table:ablation_dot_product}. As revealed by the table, the dot product is very competitive compared with a metric and it outperforms a distance metric for a couple of settings. Considering this, the dot product is still used.

\noindent \textbf{Influence of the anchor projections}. The anchor projection operation helps to summarize the information in the feature map. The ablation study is shown in \cref{table:ablation_anchor}. Considering both the accuracy performance and parameter budget, Avgpool followed by linear projection is finally used. 
\begin{table}[t]
\centering
\caption{Ablation study on similarity comparison operation.} 
\label{table:ablation_dot_product}
\vspace{-3mm}
\setlength{\tabcolsep}{1.5pt}
\scalebox{0.7}{
\begin{tabular}{c | c | c c c | c c c | c c c }
\toprule[0.1em]
\multirow{2}{*}{Test set} & \multirow{2}{*}{Metric} & \multicolumn{3}{c|}{Color DN} & \multicolumn{3}{c|}{Gray DN} & \multicolumn{3}{c}{Image SR} \\ \cline{3-11}
	& &$\sigma 15$	&$\sigma 25$	&$\sigma 50$	&$\sigma 15$	&$\sigma 25$	&$\sigma 50$	&$\times 2$	&$\times 3$	&$\times 4$	\\ \midrule
\multirow{2}{*}{\makecell{BSD68 or\\BSD100}} & Euclidean	&35.02	&32.56	&29.42	&31.84	&29.36	&26.43	&32.30	&29.19	&27.67	\\
& Dot product	&35.10	&32.64	&29.54	&31.85	&29.39	&26.44	&32.33	&29.22	&27.70	\\ \hline
\multirow{2}{*}{Urban100} & Euclidean	&34.63	&32.28	&28.94	&33.25	&30.64	&27.17	&32.76	&28.62	&26.50	\\
& Dot product	&34.77	&32.41	&29.19	&33.28	&30.75	&27.26	&32.88	&28.78	&26.67	\\
\bottomrule[0.1em]
\end{tabular}}
\vspace{-4mm}
\end{table}

\begin{table}[t]
\centering
\caption{Ablation study on anchor projection operation.} 
\label{table:ablation_anchor}
\setlength{\tabcolsep}{1.5pt}
\scalebox{0.7}{
\begin{tabular}{l | c | c }
\toprule[0.1em]
Anchor projection operation	&\# Params [m]	&PSNR on Set 5	\\ \midrule
Depthwise Conv	&3.17	&35.03	\\
Conv	&4.19	&35.03	\\
Patch merging	&3.53	&34.98	\\
Maxpool + Linear Projection	&3.12	&35.02	\\
Avgpool + Linear Projection	&3.12	&35.03	\\
\bottomrule[0.1em]
\end{tabular}}
\vspace{-4mm}
\end{table}


\section{Conclusion}
In this paper, we proposed GRL, a network with efficient and explicit hierarchical modelling capacities for image restoration. The proposed network was mainly inspired by two image properties including cross-scale similarity and anisotropic image features. Based on that, we proposed the efficient anchored stripe self-attention module for long-range dependency modelling. Then a versatile network architecture was proposed for image restoration. The proposed network can model image hierarchies in the global, regional, and local ranges. Owing to the advanced computational mechanism, the proposed network architecture achieves state-of-the-art performances for various image restoration tasks.

\vspace{1mm}
\noindent\textbf{Acknowledgements. } This work was partly supported by ETH Z\"urich General Fund (OK), Meta Reality Labs and the Alexander von Humboldt Foundation.

{\small
\bibliographystyle{ieee_fullname}
\bibliography{main}
}

\end{document}

%% file: tables/defocus_deblur_table.tex
\begin{table*}[!t]
\begin{center}
\caption{\textit{\textbf{Defocus deblurring}} results. \textbf{S:} single-image defocus deblurring. \textbf{D:} dual-pixel defocus deblurring.}
\label{table:defocus_deblurring}
\vspace{-3mm}
\setlength{\tabcolsep}{1.9pt}
\scalebox{0.7}{
\begin{tabular}{l | c | c | c | c | c | c | c | c | c | c | c | c}
\toprule[0.1em]
\multirow{2}{*}{Method} & \multicolumn{4}{c|}{\textbf{Indoor Scenes}} & \multicolumn{4}{c|}{\textbf{Outdoor Scenes}} & \multicolumn{4}{c}{\textbf{Combined}} \\ \cline{2-13}
	&PSNR$\uparrow$ & SSIM$\uparrow$ & MAE$\downarrow$ & LPIPS$\downarrow$			&PSNR$\uparrow$ & SSIM$\uparrow$ & MAE$\downarrow$ & LPIPS$\downarrow$ &PSNR$\uparrow$ & SSIM$\uparrow$ & MAE$\downarrow$ & LPIPS$\downarrow$		\\ \midrule[0.1em]
EBDB$_S$~\cite{karaali2017edge_EBDB}	&25.77	&0.772	&0.040	&0.297	&21.25	&0.599	&0.058	&0.373	&23.45	&0.683	&0.049	&0.336	\\
DMENet$_S$~\cite{lee2019deep_dmenet}	&25.50	&0.788	&0.038	&0.298	&21.43	&0.644	&0.063	&0.397	&23.41	&0.714	&0.051	&0.349	\\
JNB$_S$~\cite{shi2015just_jnb}	&26.73	&0.828	&0.031	&0.273	&21.10	&0.608	&0.064	&0.355	&23.84	&0.715	&0.048	&0.315	\\
DPDNet$_S$~\cite{abuolaim2020defocus}	&26.54	&0.816	&0.031	&0.239	&22.25	&0.682	&0.056	&0.313	&24.34	&0.747	&0.044	&0.277	\\
KPAC$_S$~\cite{son2021single_kpac}	&27.97	&0.852	&0.026	&0.182	&22.62	&0.701	&0.053	&0.269	&25.22	&0.774	&0.040	&0.227	\\
IFAN$_S$~\cite{Lee_2021_CVPRifan}	&28.11	&0.861	&0.026	&0.179	&22.76	&0.720	&0.052	&0.254	&25.37	&0.789	&0.039	&0.217	\\
Restormer$_S$~\cite{zamir2022restormer}	&\textcolor{blue}{28.87}	&\textcolor{blue}{0.882}	&\textcolor{blue}{0.025}	&\textcolor{blue}{0.145}	&\textcolor{blue}{23.24}	&\textcolor{blue}{0.743}	&\textcolor{blue}{0.050}	&\textcolor{blue}{0.209}	&\textcolor{blue}{25.98}	&\textcolor{blue}{0.811}	&\textcolor{blue}{0.038}	&\textcolor{blue}{0.178}	\\
GRL$_S$-B	&\textcolor{red}{29.06}	&\textcolor{red}{0.886}	&\textcolor{red}{0.024}	&\textcolor{red}{0.139}	&\textcolor{red}{23.45}	&\textcolor{red}{0.761}	&\textcolor{red}{0.049}	&\textcolor{red}{0.196}	&\textcolor{red}{26.18}	&\textcolor{red}{0.822}	&\textcolor{red}{0.037}	&\textcolor{red}{0.168}	\\ \midrule[0.1em]
DPDNet$_D$~\cite{abuolaim2020defocus}	&27.48	&0.849	&0.029	&0.189	&22.90	&0.726	&0.052	&0.255	&25.13	&0.786	&0.041	&0.223	\\
RDPD$_D$~\cite{abdullah2021rdpd}	&28.10	&0.843	&0.027	&0.210	&22.82	&0.704	&0.053	&0.298	&25.39	&0.772	&0.040	&0.255	\\
Uformer$_D$~\cite{wang2022uformer}	&28.23	&0.860	&0.026	&0.199	&23.10	&0.728	&0.051	&0.285	&25.65	&0.795	&0.039	&0.243	\\
IFAN$_D$~\cite{Lee_2021_CVPRifan}	&28.66	&0.868	&0.025	&0.172	&23.46	&0.743	&0.049	&0.240	&25.99	&0.804	&0.037	&0.207	\\
Restormer$_D$~\cite{zamir2022restormer}	&\textcolor{blue}{29.48}	&\textcolor{blue}{0.895}	&\textcolor{blue}{0.023}	&\textcolor{blue}{0.134}	&\textcolor{blue}{23.97}	&\textcolor{blue}{0.773}	&\textcolor{blue}{0.047}	&\textcolor{blue}{0.175}	&\textcolor{blue}{26.66}	&\textcolor{blue}{0.833}	&\textcolor{blue}{0.035}	&\textcolor{blue}{0.155}	\\
GRL$_D$-B	&\textcolor{red}{29.83}	&\textcolor{red}{0.903}	&\textcolor{red}{0.022}	&\textcolor{red}{0.114}	&\textcolor{red}{24.39}	&\textcolor{red}{0.795}	&\textcolor{red}{0.045}	&\textcolor{red}{0.150}	&\textcolor{red}{27.04}	&\textcolor{red}{0.847}	&\textcolor{red}{0.034}	&\textcolor{red}{0.133}	\\
\bottomrule[0.1em]
\end{tabular}}
\end{center}
\vspace{-8mm}
\end{table*}

%% file: tables/motion_deblur_table.tex
\begin{table}[!t]
\centering
\caption{\small \textit{\textbf{Single-image motion deblurring}} results. {GoPro} dataset \cite{nah2017deep} is used for training. 
}
\label{table:motion_deblurring}
\vspace{-3mm}
\setlength{\tabcolsep}{1.9pt}
\scalebox{0.69}{
\begin{tabular}{l | c | c | c}
\toprule[0.1em]
 & {\textbf{GoPro}  \cite{nah2017deep}} & {\textbf{HIDE}  \cite{shen2019human}} & Average \\
 \textbf{Method} & PSNR$\uparrow$ / {SSIM$\uparrow$} & PSNR$\uparrow$ / {SSIM$\uparrow$} & PSNR$\uparrow$ / {SSIM$\uparrow$} \\
\midrule[0.1em]
DeblurGAN~\cite{deblurgan}	&28.70 / 0.858		&24.51 / 0.871		&26.61 / 0.865		\\
Nah~\etal~\cite{nah2017deep}	&29.08 / 0.914		&25.73 / 0.874		&27.41 / 0.894		\\
DeblurGAN-v2~\cite{deblurganv2}	&29.55 / 0.934		&26.61 / 0.875		&28.08 / 0.905		\\
SRN~\cite{tao2018scale}	&30.26 / 0.934		&28.36 / 0.915		&29.31 / 0.925		\\
Gao \etal \cite{gao2019dynamic}	&30.90 / 0.935		&29.11 / 0.913		&30.01 / 0.924		\\
DBGAN \cite{zhang2020dbgan}	&31.10 / 0.942		&28.94 / 0.915		&30.02 / 0.929		\\
MT-RNN \cite{mtrnn2020}	&31.15 / 0.945		&29.15 / 0.918		&30.15 / 0.932		\\
DMPHN \cite{dmphn2019}	&31.20 / 0.940		&29.09 / 0.924		&30.15 / 0.932		\\
Suin \etal \cite{Maitreya2020}	&31.85 / 0.948		&29.98 / 0.930		&30.92 / 0.939		\\
SPAIR~\cite{purohit2021spatially_spair}	&32.06 / 0.953		&30.29 / 0.931		&31.18 / 0.942		\\
MIMO-UNet+~\cite{cho2021rethinking_mimo}	&32.45 / 0.957		&29.99 / 0.930		&31.22 / 0.944		\\
IPT~\cite{chen2021pre}	&32.52 / -		&- / -		&- / -		\\
MPRNet~\cite{zamir2021multi}	&32.66 / 0.959		&30.96 / 0.939		&31.81 / 0.949		\\
Restormer~\cite{zamir2022restormer}	&\textcolor{blue}{32.92} / \textcolor{blue}{0.961}		&\textcolor{blue}{31.22} / \textcolor{blue}{0.942}	&\textcolor{blue}{32.07} / \textcolor{blue}{0.952}
		\\
GRL-B (ours)	&\textcolor{red}{33.93} / \textcolor{red}{0.968}		&\textcolor{red}{31.65} / \textcolor{red}{0.947}		&\textcolor{red}{32.79} / \textcolor{red}{0.958}	\\
\bottomrule[0.1em]
\end{tabular}}
\vspace{-4mm}
\end{table}

%% file: tables/motion_deblur_realblur.tex
\begin{table}[!t]
\centering
\caption{\small \textit{\textbf{Single-image motion deblurring}} results on {RealBlur} \cite{rim2020real} dataset. The networks are trained and tested on RealBlur dataset. }
\label{table:motion_deblurring_realblur}
\vspace{-3mm}
\setlength{\tabcolsep}{1.9pt}
\scalebox{0.69}{
\begin{tabular}{l | c | c  |c}
\toprule[0.1em]
 & {\textbf{RealBlur-R} \cite{rim2020real}} & {\textbf{RealBlur-J} \cite{rim2020real}} & Average\\
 \textbf{Method} & PSNR$\uparrow$ / {SSIM$\uparrow$} & PSNR$\uparrow$ / {SSIM$\uparrow$} & PSNR$\uparrow$ / {SSIM$\uparrow$}\\
\midrule[0.1em]

DeblurGAN-v2~\cite{deblurganv2}	&36.44 / 0.935		&29.69 / 0.870		&33.07 / 0.903	\\
SRN~\cite{tao2018scale}	&38.65 / 0.965		&31.38 / 0.909		&35.02 / 0.937	\\
MPRNet~\cite{zamir2021multi}	&39.31 / 0.972		&31.76 / 0.922		&35.54 / 0.947	\\
MIMO-UNet+~\cite{cho2021rethinking_mimo}	&- / -		&32.05 / 0.921		& - / -	\\
MAXIM-3S~\cite{tu2022maxim}	&39.45 / 0.962		&\textcolor{red}{32.84} / \textcolor{red}{0.935}		&36.15 / 0.949	\\
BANet~\cite{tsai2022banet}	&39.55 / 0.971		&32.00 / 0.923		&35.78 / 0.947	\\
MSSNet~\cite{kim2022mssnet}	&39.76 / 0.972		&32.10 / 0.928		&35.93 / 0.950	\\
Stripformer~\cite{tsai2022stripformer} & \textcolor{blue}{39.84} / \textcolor{blue}{0.974} & 32.48 / 0.929 &\textcolor{blue}{36.16} / \textcolor{blue}{0.952} \\
GRL-B (ours)			&\textcolor{red}{40.20} / \textcolor{red}{0.974}		&\textcolor{blue}{32.82} / \textcolor{blue}{0.932}	 &\textcolor{red}{36.51} / \textcolor{red}{0.953} \\
\bottomrule[0.1em]
\end{tabular}}
\vspace{-4mm}
\end{table}

%% file: tables/dn_table.tex
\begin{table*}[t]
\centering
\caption{\textit{\textbf{Color and grayscale image denoising}} results. 
Model complexity and prediction accuracy are shown for better comparison. }
\label{table:denoising}
\vspace{-3mm}
\setlength{\tabcolsep}{1.5pt}
\scalebox{0.7}{
\begin{tabular}{l | r | c c c | c c c | c c c | c c c || c c c | c c c | c c c }
\toprule[0.1em]
\multirow{3}{*}{\textbf{Method}} & \multirow{3}{*}{\# \textbf{Params} [M]} & \multicolumn{12}{c||}{\textbf{Color}} & \multicolumn{9}{c}{\textbf{Grayscale}} \\ \cline{3-23}
& & \multicolumn{3}{c|}{\textbf{CBSD68}~\cite{martin2001database}} & \multicolumn{3}{c|}{\textbf{Kodak24}~\cite{franzen1999kodak}} & \multicolumn{3}{c|}{\textbf{McMaster}~\cite{zhang2011color}} & \multicolumn{3}{c||}{\textbf{Urban100}~\cite{huang2015single}}  & \multicolumn{3}{c|}{\textbf{Set12}~\cite{zhang2017beyond}} & \multicolumn{3}{c|}{\textbf{BSD68}~\cite{martin2001database}} & \multicolumn{3}{c}{\textbf{Urban100}~\cite{huang2015single}} \\
        &  & $\sigma$$=$$15$ & $\sigma$$=$$25$ & $\sigma$$=$$50$ & $\sigma$$=$$15$ & $\sigma$$=$$25$ & $\sigma$$=$$50$ & $\sigma$$=$$15$ & $\sigma$$=$$25$ & $\sigma$$=$$50$ & $\sigma$$=$$15$ & $\sigma$$=$$25$ & $\sigma$$=$$50$ & $\sigma$$=$$15$ & $\sigma$$=$$25$ & $\sigma$$=$$50$ & $\sigma$$=$$15$ & $\sigma$$=$$25$ & $\sigma$$=$$50$ & $\sigma$$=$$15$ & $\sigma$$=$$25$ & $\sigma$$=$$50$ \\ \midrule
DnCNN~\cite{kiku2016beyond}	&\textcolor{red}{0.56}	&33.90	&31.24	&27.95	&34.60	&32.14	&28.95	&33.45	&31.52	&28.62	&32.98	&30.81	&27.59				&32.86	&30.44	&27.18	&31.73	&29.23	&26.23	&32.64	&29.95	&26.26	\\
RNAN~\cite{zhang2019residual}	&8.96	&-	&-	&28.27	&-	&-	&29.58	&-	&-	&29.72	&-	&-	&29.08				&-	&-	&27.70	&-	&-	&26.48	&-	&-	&27.65	\\
IPT~\cite{chen2021pre}	&115.33	&-	&-	&28.39	&-	&-	&29.64	&-	&-	&29.98	&-	&-	&29.71				&-	&-	&-	&-	&-	&-	&-	&-	&-	\\
EDT-B~\cite{li2021efficient}	&11.48	&34.39	&31.76	&28.56	&35.37	&32.94	&29.87	&\textcolor{blue}{35.61}	&\textcolor{blue}{33.34}	&30.25	&35.22	&\textcolor{blue}{33.07}	&\textcolor{blue}{30.16}				&-	&-	&-	&-	&-	&-	&-	&-	&-	\\
DRUNet~\cite{zhang2021plug}	&32.64	&34.30	&31.69	&28.51	&35.31	&32.89	&29.86	&35.40	&33.14	&30.08	&34.81	&32.60	&29.61				&33.25	&30.94	&27.90	&31.91	&29.48	&26.59	&33.44	&31.11	&27.96	\\
SwinIR~\cite{liang2021swinir}	&11.75	&\textcolor{blue}{34.42}	&31.78	&28.56	&35.34	&32.89	&29.79	&\textcolor{blue}{35.61}	&33.20	&30.22	&35.13	&32.90	&29.82				&33.36	&31.01	&27.91	&\textcolor{blue}{31.97}	&29.50	&26.58	&33.70	&31.30	&27.98	\\
Restormer ~\cite{zamir2022restormer}	&26.13	&34.40	&\textcolor{blue}{31.79}	&\textcolor{blue}{28.60}	&\textcolor{red}{35.47}	&\textcolor{red}{33.04}	&\textcolor{red}{30.01}	&\textcolor{blue}{35.61}	&\textcolor{blue}{33.34}	&\textcolor{blue}{30.30}	&35.13	&32.96	&30.02				&\textcolor{blue}{33.42}	&\textcolor{blue}{31.08}	&\textcolor{blue}{28.00}	&31.96	&\textcolor{blue}{29.52}	&\textcolor{red}{26.62}	&33.79	&31.46	&\textcolor{blue}{28.29}	\\
GRL-T	&\textcolor{blue}{0.88}	&34.30	&31.66	&28.45	&35.24	&32.78	&29.67	&35.49	&33.18	&30.06	&35.08	&32.84	&29.78				&33.29	&30.92	&27.78	&31.90	&29.43	&26.49	&33.66	&31.23	&27.89	\\
GRL-S	&3.12	&34.36	&31.72	&28.51	&35.32	&32.88	&29.77	&35.59	&33.29	&30.18	&\textcolor{blue}{35.24}	&\textcolor{blue}{33.07}	&30.09				&33.36	&31.02	&27.91	&31.93	&29.47	&26.54	&\textcolor{blue}{33.84}	&\textcolor{blue}{31.49}	&28.24	\\
GRL-B	&19.81	&\textcolor{red}{34.45}	&\textcolor{red}{31.82}	&\textcolor{red}{28.62}	&\textcolor{blue}{35.43}	&\textcolor{blue}{33.02}	&\textcolor{blue}{29.93}	&\textcolor{red}{35.73}	&\textcolor{red}{33.46}	&\textcolor{red}{30.36}	&\textcolor{red}{35.54}	&\textcolor{red}{33.35}	&\textcolor{red}{30.46}				&\textcolor{red}{33.47}	&\textcolor{red}{31.12}	&\textcolor{red}{28.03}	&\textcolor{red}{32.00}	&\textcolor{red}{29.54}	&\textcolor{blue}{26.60}	&\textcolor{red}{34.09}	&\textcolor{red}{31.80}	&\textcolor{red}{28.59}	\\		
\bottomrule[0.1em]
\end{tabular}}
\end{table*}

%% file: tables/sr_table.tex
\begin{table*}[t]\scriptsize
\center
\begin{center}
\caption{\textbf{\textit{Classical image SR}} results. Results of both lightweight models and accurate models are summarized.}
\vspace{-3mm}
\label{tab:sr_results}
\begin{tabular}{l|c|r|c|c|c|c|c|c|c|c|c|c}
\toprule[0.1em]
\multirow{2}{*}{\textbf{Method}} & \multirow{2}{*}{\textbf{Scale}} & \multirow{2}{*}{\# \textbf{Params} [M]} &  \multicolumn{2}{c|}{\textbf{Set5}~\cite{bevilacqua2012low}} &  \multicolumn{2}{c|}{\textbf{Set14}~\cite{zeyde2010single}} &  \multicolumn{2}{c|}{\textbf{BSD100}~\cite{martin2001database}} &  \multicolumn{2}{c|}{\textbf{Urban100}~\cite{huang2015single}} &  \multicolumn{2}{c}{\textbf{Manga109}~\cite{matsui2017sketch}}  
\\
\cline{4-13}
&  &  & PSNR$\uparrow$ & SSIM$\uparrow$ & PSNR$\uparrow$ & SSIM$\uparrow$ & PSNR$\uparrow$ & SSIM$\uparrow$ & PSNR$\uparrow$ & SSIM$\uparrow$ & PSNR$\uparrow$ & SSIM$\uparrow$ 
\\
\midrule[0.1em]
RCAN~\cite{zhang2018rcan}&	×2&	15.44&	38.27&	0.9614&	34.12&	0.9216&	32.41&	0.9027&	33.34&	0.9384&	39.44&	0.9786\\
SAN~\cite{dai2019SAN}&	×2&	15.71&	38.31&	0.9620&	34.07&	0.9213&	32.42&	0.9028&	33.10&	0.9370&	39.32&	0.9792\\
HAN~\cite{niu2020HAN}&	×2&	63.61&	38.27&	0.9614&	34.16&	0.9217&	32.41&	0.9027&	33.35&	0.9385&	39.46&	0.9785\\
IPT~\cite{chen2021pre}&	×2&	115.48&	38.37&	-&	34.43&	-&	32.48&	-&	33.76&	-&	-&	-\\ \hline
SwinIR~\cite{liang2021swinir}&	×2&	\textcolor{red}{0.88}&	38.14&	0.9611&	33.86&	0.9206&	32.31&	0.9012&	32.76&	0.9340&	39.12&	0.9783\\
SwinIR~\cite{liang2021swinir}&	×2&	11.75&	38.42&	0.9623&	34.46&	0.9250&	32.53&	0.9041&	33.81&	0.9427&	39.92&	0.9797\\  
EDT~\cite{li2021efficient}&	×2&	0.92&	38.23&	0.9615&	33.99&	0.9209&	32.37&	0.9021&	32.98&	0.9362&	39.45&	0.9789\\
EDT~\cite{li2021efficient}&	×2&	11.48&	\textcolor{blue}{38.63}&	\textcolor{blue}{0.9632}&	\textcolor{blue}{34.80}&	0.9273&	\textcolor{blue}{32.62}&	0.9052&	34.27&	0.9456&	\textcolor{blue}{40.37}&	\textcolor{blue}{0.9811}\\ 
GRL-T (ours) &	×2&	\textcolor{blue}{0.89}&	38.27&	0.9627&	34.21&	0.9258&	32.42&	0.9056&	33.60&	0.9411&	39.61&	0.9790\\
GRL-S (ours) &	×2&	3.34&	38.37&	\textcolor{blue}{0.9632}&	34.64&	\textcolor{blue}{0.9280}&	32.52&	\textcolor{blue}{0.9069}&	\textcolor{blue}{34.36}&	\textcolor{blue}{0.9463}&	39.84&	0.9793\\
GRL-B (ours) &	×2&	20.05&	\textcolor{red}{38.67}&	\textcolor{red}{0.9647}&	\textcolor{red}{35.08}&	\textcolor{red}{0.9303}&	\textcolor{red}{32.67}&	\textcolor{red}{0.9087}&	\textcolor{red}{35.06}&	\textcolor{red}{0.9505}&	\textcolor{red}{40.67}&	\textcolor{red}{0.9818}\\ \midrule[0.1em]
RCAN~\cite{zhang2018rcan}&	×4&	15.59&	32.63&	0.9002&	28.87&	0.7889&	27.77&	0.7436&	26.82&	0.8087&	31.22&	0.9173\\
SAN~\cite{dai2019SAN}&	×4&	15.86&	32.64&	0.9003&	28.92&	0.7888&	27.78&	0.7436&	26.79&	0.8068&	31.18&	0.9169\\
HAN~\cite{niu2020HAN}&	×4&	64.20&	32.64&	0.9002&	28.90&	0.7890&	27.80&	0.7442&	26.85&	0.8094&	31.42&	0.9177\\
IPT~\cite{chen2021pre}&	×4&	115.63&	32.64&	-&	29.01&	-&	27.82&	-&	27.26&	-&	-&	-\\ \hline
SwinIR~\cite{liang2021swinir}&	×4&	\textcolor{red}{0.90}&	32.44&	0.8976&	28.77&	0.7858&	27.69&	0.7406&	26.47&	0.7980&	30.92&	0.9151\\
SwinIR~\cite{liang2021swinir}&	×4&	11.90&	32.92&	0.9044&	29.09&	0.7950&	27.92&	0.7489&	27.45&	0.8254&	32.03&	0.9260\\  
EDT~\cite{li2021efficient}&	×4&	0.92&	32.53&	0.8991&	28.88&	0.7882&	27.76&	0.7433&	26.71&	0.8051&	31.35&	0.9180\\
EDT~\cite{li2021efficient}&	×4&	11.63&	\textcolor{blue}{33.06}&	0.9055&	\textcolor{blue}{29.23}&	0.7971&	\textcolor{blue}{27.99}&	0.7510&	27.75&	0.8317&	\textcolor{blue}{32.39}&	\textcolor{blue}{0.9283}\\ 
GRL-T (ours) &	×4&	\textcolor{blue}{0.91}&	32.56&	0.9029&	28.93&	0.7961&	27.77&	0.7523&	27.15&	0.8185&	31.57&	0.9219\\
GRL-S (ours) &	×4&	3.49&	32.76&	\textcolor{blue}{0.9058}&	29.10&	\textcolor{blue}{0.8007}&	27.90&	\textcolor{blue}{0.7568}&	\textcolor{blue}{27.90}&	\textcolor{blue}{0.8357}&	32.11&	0.9267\\
GRL-B (ours) &	×4&	20.20&	\textcolor{red}{33.10}&	\textcolor{red}{0.9094}&	\textcolor{red}{29.37}&	\textcolor{red}{0.8058}&	\textcolor{red}{28.01}&	\textcolor{red}{0.7611}&	\textcolor{red}{28.53}&	\textcolor{red}{0.8504}&	\textcolor{red}{32.77}&	\textcolor{red}{0.9325}\\ \bottomrule[0.1em]
\end{tabular}
\end{center}
\vspace{-4mm}
\end{table*}

%% file: tables/jpeg_table.tex
\begin{table*}[t]
\parbox{.62\linewidth}{
\centering
\caption{\textit{\textbf{Grayscale image JPEG compression artifact removal}} results. As a comparison metric, the parameter count of FBCNN~\cite{jiang2021FBCNN} GRL-S are 71.92M and 3.12M.} 
\label{table:jpeg_car}
\vspace{-3mm}
\setlength{\tabcolsep}{1.5pt}
\scalebox{0.7}{
\begin{tabular}{c | c| c c | c c | c c | c c | c c | c c | c c }
\toprule[0.1em]
\multirow{2}{*}{Set} & \multirow{2}{*}{QF} & \multicolumn{2}{c|}{JPEG}  & \multicolumn{2}{c|}{\makecell{DnCNN  \cite{zhang2017beyond}}} & \multicolumn{2}{c|}{\makecell{DCSC \cite{fu2019jpeg}}} & \multicolumn{2}{c|}{\makecell{QGAC \cite{ehrlich2020quantization}}} & \multicolumn{2}{c|}{\makecell{MWCNN \cite{liu2018MWCNN}}} & \multicolumn{2}{c|}{\makecell{FBCNN \cite{jiang2021FBCNN}}} & \multicolumn{2}{c}{GRL-S}  \\
& & PSNR & SSIM & PSNR & SSIM & PSNR & SSIM & PSNR & SSIM & PSNR & SSIM & PSNR & SSIM & PSNR & SSIM  \\
\midrule[0.1em]
\parbox[t]{0.8cm}{\multirow{3}{*}{\rotatebox[origin=c]{90}{\makecell{Classic5 \\ \cite{foi2007Classic5}}}}}
    &10	&27.82	&0.760			&29.40	&0.803		&29.62	&0.810		&29.84	&0.812		&30.01	&0.820		&30.12	&0.822		&\textcolor{red}{30.20}	&\textcolor{red}{0.829}	\\
    &20	&30.12	&0.834			&31.63	&0.861		&31.81	&0.864		&31.98	&0.869		&32.16	&0.870		&32.31	&0.872		&\textcolor{red}{32.49}	&\textcolor{red}{0.878}	\\
    &30	&31.48	&0.867			&32.91	&0.886		&33.06	&0.888		&33.22	&0.892		&33.43	&0.893		&33.54	&0.894		&\textcolor{red}{33.72}	&\textcolor{red}{0.899}	\\
    &40	&32.43	&0.885			&33.77	&0.900		&33.87	&0.902		&34.05	&0.905		&34.27	&0.906		&34.35	&0.907		&\textcolor{red}{34.53}	&\textcolor{red}{0.911}	\\ \midrule[0.1em]
\parbox[t]{0.8cm}{\multirow{3}{*}{\rotatebox[origin=c]{90}{\makecell{BSD500 \\ \cite{martin2001database}}}}}                                                                    
    &10	&27.80	&0.768			&29.21	&0.809		&29.32	&0.813		&29.46	&0.821		&29.61	&0.820		&29.67	&0.821		&\textcolor{red}{29.74}	&\textcolor{red}{0.823}	\\
    &20	&30.05	&0.849			&31.53	&0.878		&31.63	&0.880		&31.73	&0.884		&31.92	&\textcolor{red}{0.885}		&32.00	&\textcolor{red}{0.885}		&\textcolor{red}{32.05}	&\textcolor{red}{0.885}	\\
    &30	&31.37	&0.884			&32.90	&0.907		&32.99	&0.908		&33.07	&0.912		&33.30	&0.912		&33.37	&\textcolor{red}{0.913}		&\textcolor{red}{33.43}	&0.912	\\
    &40	&32.30	&0.903			&33.85	&0.923		&33.92	&0.924		&34.01	&0.927		&34.27	&\textcolor{red}{0.928}		&34.33	&\textcolor{red}{0.928}		&\textcolor{red}{34.38}	&\textcolor{red}{0.928}	\\
\bottomrule[0.1em]
\end{tabular}}
}
\hfill
\parbox{.38\linewidth}{
\centering
\caption{\textit{\textbf{Color image JPEG compression artifact removal} results.}}
\label{table:colordenoising}
\vspace{-3mm}
\setlength{\tabcolsep}{1.5pt}
\scalebox{0.7}{
\begin{tabular}{c | c | c c | c c | c c | c c}
\toprule[0.1em]
\multirow{2}{*}{Set} & \multirow{2}{*}{QF} & \multicolumn{2}{c|}{JPEG}  & \multicolumn{2}{c|}{\makecell{QGAC \cite{ehrlich2020quantization}}} & \multicolumn{2}{c|}{\makecell{FBCNN \cite{jiang2021FBCNN}}} & \multicolumn{2}{c}{GRL-S} \\ 
& & PSNR & SSIM & PSNR & SSIM & PSNR & SSIM & PSNR & SSIM  \\
\midrule[0.1em]
\parbox[t]{0.8cm}{\multirow{3}{*}{\rotatebox[origin=c]{90}{\makecell{LIVE1 \\ \cite{sheikh2005live}}}}}
        &10	&25.69	&0.743		&27.62	&0.804		&27.77	&0.803		&\textcolor{red}{28.13}	&\textcolor{red}{0.814}	\\
	&20	&28.06	&0.826		&29.88	&0.868		&30.11	&0.868		&\textcolor{red}{30.49}	&\textcolor{red}{0.878}	\\
	&30	&29.37	&0.861		&31.17	&0.896		&31.43	&0.897		&\textcolor{red}{31.85}	&\textcolor{red}{0.905}	\\
	&40	&30.28	&0.882		&32.05	&0.912		&32.34	&0.913		&\textcolor{red}{32.79}	&\textcolor{red}{0.920}	\\ \midrule[0.1em]
\parbox[t]{0.8cm}{\multirow{3}{*}{\rotatebox[origin=c]{90}{\makecell{BSD500 \\ \cite{martin2001database}}}}}
        &10	&25.84	&0.741		&27.74	&0.802		&27.85	&0.799		&\textcolor{red}{28.26}	&\textcolor{red}{0.808}	\\
	&20	&28.21	&0.827		&30.01	&0.869		&30.14	&0.867		&\textcolor{red}{30.57}	&\textcolor{red}{0.875}	\\
	&30	&29.57	&0.865		&31.33	&0.898		&31.45	&0.897		&\textcolor{red}{31.92}	&\textcolor{red}{0.903}	\\
	&40	&30.52	&0.887		&32.25	&0.915		&32.36	&0.913		&\textcolor{red}{32.86}	&\textcolor{red}{0.919}	\\
\bottomrule[0.1em]
\end{tabular}}
}
\vspace{-4mm}
\end{table*}

%% file: tables/demosaicking_table.tex
\begin{table}[!t]
\begin{center}
\caption{\textit{\textbf{Image demosaicking}} results.}
\label{table:demosaicking}
\vspace{-3mm}
\setlength{\tabcolsep}{1.9pt}
\scalebox{0.7}{
\begin{tabular}{l | c c c c c c c c }
\toprule
Datasets	&Matlab	&\makecell{DDR \\ \cite{wu2016demosaicing}}	&\makecell{DeepJoint \\ \cite{gharbi2016deep}}	&\makecell{MMNet \\ \cite{kokkinos2019iterative}}	&\makecell{RLDD \\ \cite{guo2020residual}}	&\makecell{RNAN \\ \cite{zhang2019residual}}	&\makecell{DRUNet \\ \cite{zhang2021plug}}	&\makecell{GRL-S \\ (ours)}	\\ \hline
Kodak~\cite{franzen1999kodak}	&35.78	&41.11	&42.00	&40.19	&42.49	&\textcolor{blue}{43.16}	&42.68	&\textcolor{red}{43.57}	\\
McMaster~\cite{zhang2011color}	&34.43	&37.12	&39.14	&37.09	&39.25	&\textcolor{blue}{39.70}	&39.39	&\textcolor{red}{40.22}	\\
\bottomrule[0.1em]
\end{tabular}}
\end{center}
\vspace{-4mm}
\end{table}